\pdfoutput=1
\documentclass[11pt]{article}

\usepackage[utf8]{inputenc}
\usepackage[T1]{fontenc}
\usepackage[dvipsnames]{xcolor}
\definecolor{hunyuanblue}{HTML}{1E4A8F}
\definecolor{abstractbg}{HTML}{F0F7FC}

\usepackage{float}
\usepackage{hunyuan}

\usepackage{inconsolata}
\usepackage{booktabs}
\usepackage{array}
\usepackage{multirow}
\usepackage{enumitem}
\usepackage{microtype}
\usepackage[most]{tcolorbox}
\usepackage{fancyhdr}
\usepackage{xurl}
\usepackage{placeins}

\newtcolorbox{promptbox}[1]{%
  enhanced, breakable,
  colback=abstractbg, colframe=hunyuanblue!40, boxrule=0.5pt, arc=2mm,
  colbacktitle=hunyuanblue!80, fonttitle=\bfseries\small\color{white},
  title={#1},
  attach boxed title to top left={yshift=-2.6mm, xshift=3.5mm},
  boxed title style={arc=1mm, boxrule=0pt},
  boxsep=4pt, left=6pt, right=6pt, top=4.2mm, bottom=4pt,
  before skip=10pt, after skip=12pt,
  fontupper=\small\ttfamily, halign=left,
  before upper={\setlength{\parskip}{0.5em}}
}

\hypersetup{colorlinks=true, linkcolor=hunyuanblue, citecolor=hunyuanblue, urlcolor=hunyuanblue,
  pdftitle={AFCS: Atomic-Fact-Conditioned Synthesis for Factual Knowledge Pretraining},
  pdfauthor={Yunhao Wang, Feng Zhang, Tinghao Yu, Xing Wu, Jiang Zhou}}

\newlength{\fulltextwidth}
\makeatletter
\long\def\@makecaption#1#2{%
  \vskip\abovecaptionskip
  \sbox\@tempboxa{#1: #2}%
  \ifdim \wd\@tempboxa >\hsize
    {\ifdim\hsize<\fulltextwidth\raggedright\fi #1: #2\par}%
  \else
    \global\@minipagefalse
    \hb@xt@\hsize{\hfil\box\@tempboxa\hfil}%
  \fi
  \vskip\belowcaptionskip}
\def\section{\@startsiction{section}{1}{\z@}{-0.24in}{0.10in}
             {\large\bf\raggedright\color{hunyuanblue}}}
\def\subsection{\@startsection{subsection}{2}{\z@}{-0.20in}{0.08in}
                {\normalsize\bf\raggedright\color{hunyuanblue}\nobreak}}
\makeatother

\setlist[itemize]{leftmargin=1.35em,itemsep=0.2em,topsep=0.25em}
\setlist[enumerate]{leftmargin=1.55em,itemsep=0.2em,topsep=0.25em}

\fancypagestyle{plain}{\fancyhf{}\fancyfoot[C]{\thepage}}
\fancypagestyle{firststyle}{%
  \fancyhf{}%
  \fancyhead[L]{\IfFileExists{hunyuan-logo.png}{\raisebox{-17mm}[0pt][0pt]{\includegraphics[height=10mm]{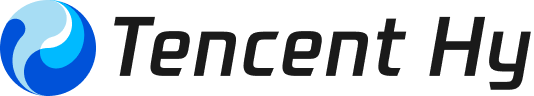}}}{}}%
  \fancyfoot[C]{\thepage}%
}

\begin{document}
\thispagestyle{firststyle}
\vspace*{0.25cm}
{\color{hunyuanblue}\hrule height 0.6pt}
\vskip 6mm
\begin{center}
{\LARGE\bfseries\color{black}
AFCS: Atomic-Fact-Conditioned Synthesis for Factual Knowledge Pretraining\par}
\end{center}
\vskip 3mm
{\color{hunyuanblue}\hrule height 0.6pt}
\vskip 6mm

\begin{center}
\textbf{
Yunhao Wang,
Feng Zhang,
Tinghao Yu,
Xing Wu,
Jiang Zhou
}\\[4pt]
Hunyuan Team, Tencent\\[2pt]
\texttt{\{luciuswang,jayzhang\}@tencent.com}
\end{center}

\vskip 5mm

\begin{tcolorbox}[colframe=abstractbg, colback=abstractbg, boxrule=0pt, arc=2mm, enhanced, top=11pt, bottom=11pt, left=14pt, right=14pt, width=\textwidth]
\textbf{Abstract.}\quad
Synthetic data is increasingly used in language-model pretraining. Existing knowledge-corpus synthesis pipelines, including WRAP and Nemotron-CC, usually take a source passage or document segment as input and generate rewritten or QA-style samples from that text. This can improve surface form and prompt diversity, but it does not explicitly require fine-grained source facts---especially secondary entities, list/table entries, and precise temporal attributes---to appear in the synthesized corpus. We propose \textbf{AFCS} (\textbf{A}tomic-\textbf{F}act-\textbf{C}onditioned \textbf{S}ynthesis), a pipeline that first extracts source-supported atomic facts and then uses each retained fact to drive generation. AFCS uses structure-aware extraction over the document tree, source-grounded verification, local context expansion, and multi-format generation. In continued-pretraining experiments with OLMo-3-7B and a 32B base on the same FineWiki source corpus, AFCS outperforms WRAP and Nemotron-CC on SimpleQA, SimpleQA Verified, PopQA, and HotpotQA. Granularity ablations show that fact-level synthesis improves over chunk-level, paragraph-level, and entity-level synthesis; and a 100B-token mid-training integration improves factual benchmarks and raises the 12-task general benchmark average by $+1.09$.
\end{tcolorbox}

\section{Introduction}
\label{sec:intro}

Synthetic data has become a common component of recent language-model pretraining recipes. WRAP and Nemotron-CC transform source passages or document segments into rewritten, QA-style, and knowledge-focused outputs \citep{maini2024rephrasing,su2025nemotron}. Broader pretraining recipes also incorporate synthetic textbooks, question--answer pairs, instructions, exercises, and structured reasoning data \citep{abdin2024phi4technicalreport,yang2025qwen3}. Together, these approaches expose knowledge through diverse textual forms during pretraining.

Most knowledge-corpus synthesis methods take a source text unit as input. WRAP rewrites a source passage into multiple styles \citep{maini2024rephrasing}, while Nemotron-CC applies a multi-prompt mixture to document segments \citep{su2025nemotron}. These methods differ in output format, but they share a common interface: a passage, segment, or document is given to the generator, which then produces synthetic samples from that input.

This work studies a fact-coverage issue that arises under this interface. Source-conditioned generation often preserves salient entities and headline events, but may omit secondary entities, list entries, table rows, and fine-grained dates. We treat this as a data-construction problem rather than as a ground-truth coverage metric: diagnostic extraction and matching are used to identify recurring omission patterns, while the continued-pretraining experiments provide the main evidence for whether the resulting synthetic data improves model behavior.

We propose \textbf{Atomic-Fact-Conditioned Synthesis (AFCS)}. AFCS changes the driver of data synthesis. Instead of using a source text unit alone as the generation prompt, AFCS first extracts source-supported atomic facts, expands the local context when needed, and generates samples conditioned on retained facts selected for synthesis. AFCS targets explicit fact coverage in the synthetic corpus; its effect on model behavior is evaluated through downstream continued-pretraining experiments.

Our contributions are as follows:
\begin{enumerate}[nosep,leftmargin=*]
\item We show that source-conditioned knowledge-corpus synthesis can omit many fine-grained facts that are present in the source, especially for secondary entities, list/table entries, and fine-grained temporal attributes.
\item We introduce AFCS as a coverage-oriented data-construction interface that turns extracted, source-supported assertions into explicit generation targets through structure-aware extraction, source-grounded verification, local context expansion, and multi-format generation.
\item We evaluate AFCS on OLMo-3-7B \citep{olmo2025olmo3} and a 32B base using the same FineWiki source corpus \citep{penedo2025finewiki} as the baselines. AFCS improves middle- and long-tail factual recall on PopQA.
\item Through granularity and mechanism ablations, we show that fact-level conditioning outperforms coarser units; targeted mechanism contrasts yield results consistent with benefits from broader fact coverage, repeated realizations per fact, and multi-format generation.
\end{enumerate}

\section{Related Work}
\label{sec:related}

\paragraph{Synthetic-data pretraining and web-corpus refinement.} Continued pretraining on domain-specific corpora is a standard way to adapt language models \citep{gururangan2020don}. Recent pretraining recipes increasingly use synthetic data to improve the presentation of raw text: WRAP paraphrasing rewrites web passages into cleaner QA or expository forms \citep{maini2024rephrasing}, Nemotron-CC applies a multi-prompt mixture to Common Crawl segments \citep{su2025nemotron}, and other systems use genre reformulation and active reading \citep{hu2025maga,lin2025activereading}. Web-corpus construction work such as FineWeb and DCLM studies filtering, deduplication, and source-quality trade-offs at scale \citep{penedo2024fineweb,li2024dclm}. These works focus on source quality, surface form, and corpus-scale construction. AFCS addresses a narrower interface question: for a fixed source corpus, how should source facts be selected as synthetic-data targets before they are realized in text?

\paragraph{Source-conditioned rewriting versus fact-conditioned synthesis.} Most existing synthesis pipelines condition generation on a passage, document segment, or whole article. Prompt diversity can broaden the forms in which a document is studied, and structure-preserving paraphrasing can maintain label fidelity, but the generator still decides which source facts to mention. Active Reading similarly broadens document-conditioned study through model-proposed strategies including paraphrasing, knowledge linking, active recall, and analogical reasoning \citep{lin2025activereading}. These approaches increase the diversity of views or actions applied to a source document, but they do not make source-fact selection explicit before generation. This creates a fact-selection bottleneck: salient entities and headline events are likely to be preserved, while secondary entities, list entries, table rows, and precise dates may be omitted. AFCS changes this interface by extracting source-grounded facts before generation and assigning generation attempts to those facts. The distinction is not the use of QA or rewriting formats, but whether fact selection remains implicit in a source-text prompt or is made explicit before surface realization.

\paragraph{Entity-, relation-, and graph-conditioned synthesis.} Web structure can provide intermediate signals for pretraining: Hyperlink-induced pretraining \citep{zhou2022hyperlink} uses hyperlink topology to construct pseudo query--passage pairs for retrieval. For synthetic continued pretraining, EntiGraph \citep{yang2024synthetic} extracts entities and synthesizes from entity pairs or triples; Synthesize-on-Graph (SoG) \citep{ma2025sog} builds a context graph and uses graph traversal followed by secondary sampling to construct cross-document generation contexts; Synthetic Bootstrapped Pretraining \citep{yang2025sbp} learns inter-document relations and synthesizes new corpora from those relations. Structured methods based on entities, attributes, relations, or graphs can represent many of the same facts targeted by AFCS, especially when qualifiers, reification, or schema extensions are allowed. We therefore do not view the distinction as one of theoretical representational capacity. Rather, AFCS makes fact selection explicit and source-local: each extracted, source-supported assertion is treated as a generation target, with its target value and evidence span available for realization. The anchor of such an assertion need not be a canonical entity; it can also be a table row, list item, section context, date, numerical value, or descriptive span. Several AFCS assertions encode negation, evidential absence, conditionality, or full-clause values that are not naturally represented as a single canonical entity--relation edge (Appendix~\ref{app:noncanonical}). Entity- or graph-conditioned methods typically provide entities, attributes, profiles, relation edges, or graph paths as synthesis context, leaving the generator or graph-walk procedure to determine which facts are ultimately realized in text. Our matched-budget entity-level ablation provides one comparison with a coarser conditioning unit and yields lower scores than conditioning on explicit fact-value assertions in our setting. In different knowledge-injection settings, Mecklenburg et al.'s fact-based scaling pipeline \citep{mecklenburg2024injecting} constructs QA pairs from extracted atomic facts, while Knowledge-Instruct \citep{ovadia2025knowledge} constructs instruction data from entity-anchored facts.

\paragraph{Factual knowledge evaluation.} SimpleQA evaluates short, fact-seeking questions with a single verifiable answer \citep{wei2024simpleqa}. We use factual QA benchmarks to compare model behavior before and after continued pretraining. Broad benchmarks such as MMLU measure general knowledge and reasoning across many domains \citep{hendrycks2021measuring}. In this work, SimpleQA and SimpleQA Verified \citep{wei2024simpleqa,haas2025simpleqaverified}, PopQA \citep{mallen2023trust}, and HotpotQA \citep{yang2018hotpotqa} measure factual QA behavior, while a separate source-fact coverage diagnostic evaluates the synthetic corpus and the 12-task mid-training suite checks broader capability compatibility. This separation is important because explicit source-fact coverage does not automatically imply that every covered fact is stored or retrieved by the trained model.

\paragraph{Knowledge editing and retrieval.} ROME \citep{meng2022locating} edits individual factual associations, while MEMIT \citep{meng2023massediting} directly edits many factual memories in model parameters; neither is a corpus-scale pretraining method. RAG \citep{lewis2020rag} combines parametric generation with an external non-parametric memory at inference time, while REPLUG \citep{shi2024replug} motivates retrieval as access to the long tail of knowledge that a parametric model may not represent. AFCS instead changes the pretraining corpus, while retrieval-augmented methods remain complementary for deployment-time access to external evidence.

\section{Problem Formulation: Source-Fact Coverage}
\label{sec:coverage}

\subsection{Atomic Facts}
\label{sec:atomic-facts}

Let $\mathcal{D}=\{d_1,\dots,d_N\}$ be a knowledge corpus. For a source unit $d$, let $\mathcal{F}(d)=\{f_1,\dots,f_{n_d}\}$ denote the atomic facts expressed in that unit. We use \emph{atomic fact} to mean a minimal source-supported assertion that can be verified against a local evidence span. The unit is not required to be centered on a canonical entity or to correspond to a canonical entity--relation edge. Some records are anchored by named entities, while others are anchored by a table row, list item, section context, date, numerical value, or descriptive span. In implementation, records are stored in a subject--predicate--value-like form for convenience, but the subject can be a local anchor rather than a canonical entity, and the predicate or value may contain free-form natural language. Each record contains a target value and a source span that supports it. Appendix~\ref{app:noncanonical} gives examples of assertions that use this notation but are not intended as canonical KG triples.

A synthesis algorithm maps either a source unit or an extracted fact set to training samples. We say a sample covers a fact when it explicitly states the fact value with enough local context to identify the relevant entity or attribute. This notion describes the source-fact selection problem; the paper's main model-performance claims are based on downstream continued-pretraining results, with coverage reported separately as a data-side diagnostic.

\subsection{Source-Conditioned Synthesis}
\label{sec:source_conditioned}

Prior synthesis methods can be organized by their conditioning unit. WRAP \citep{maini2024rephrasing} produces prompted rewrites from source passages or paragraph-level examples, including QA-style and stylistic paraphrase outputs. Nemotron-CC \citep{su2025nemotron} defines five prompt types for document segments with prompt-specific token limits, including knowledge extraction, distillation, diverse QA pairs, knowledge lists, and Wikipedia-style rewriting.

These pipelines can produce fluent and diverse text, but they do not directly specify which fine-grained source facts must be expressed. The generator receives a source unit and produces synthetic samples from it. We call this family \emph{source-conditioned} synthesis. AFCS instead uses \emph{atomic-fact-conditioned} synthesis: an extractor first identifies atomic facts, and generation is conditioned on retained facts selected for synthesis, with source context available for realization.

\subsection{Omission Patterns}
\label{sec:funnel}

We separate missing source evidence from omissions introduced by synthesis; the SimpleQA Verified source-availability funnel is reported in Appendix~\ref{app:source-availability}. The main text focuses on source facts that are present in the corpus but are not explicitly realized by source-conditioned synthesis.

Author inspection of 200 uncovered facts from the source-conditioned pipeline shows three recurring omission patterns. First, secondary entities---including people, organizations, and locations mentioned relatively infrequently within an article---are often omitted. Second, list- and table-embedded facts are under-covered: a QA generator may ask about an award as a whole while omitting individual year--recipient rows. Third, temporal facts are often coarsened, for example from month/day precision to year-level statements.

\begin{figure}[t]
\centering
\includegraphics[width=\linewidth]{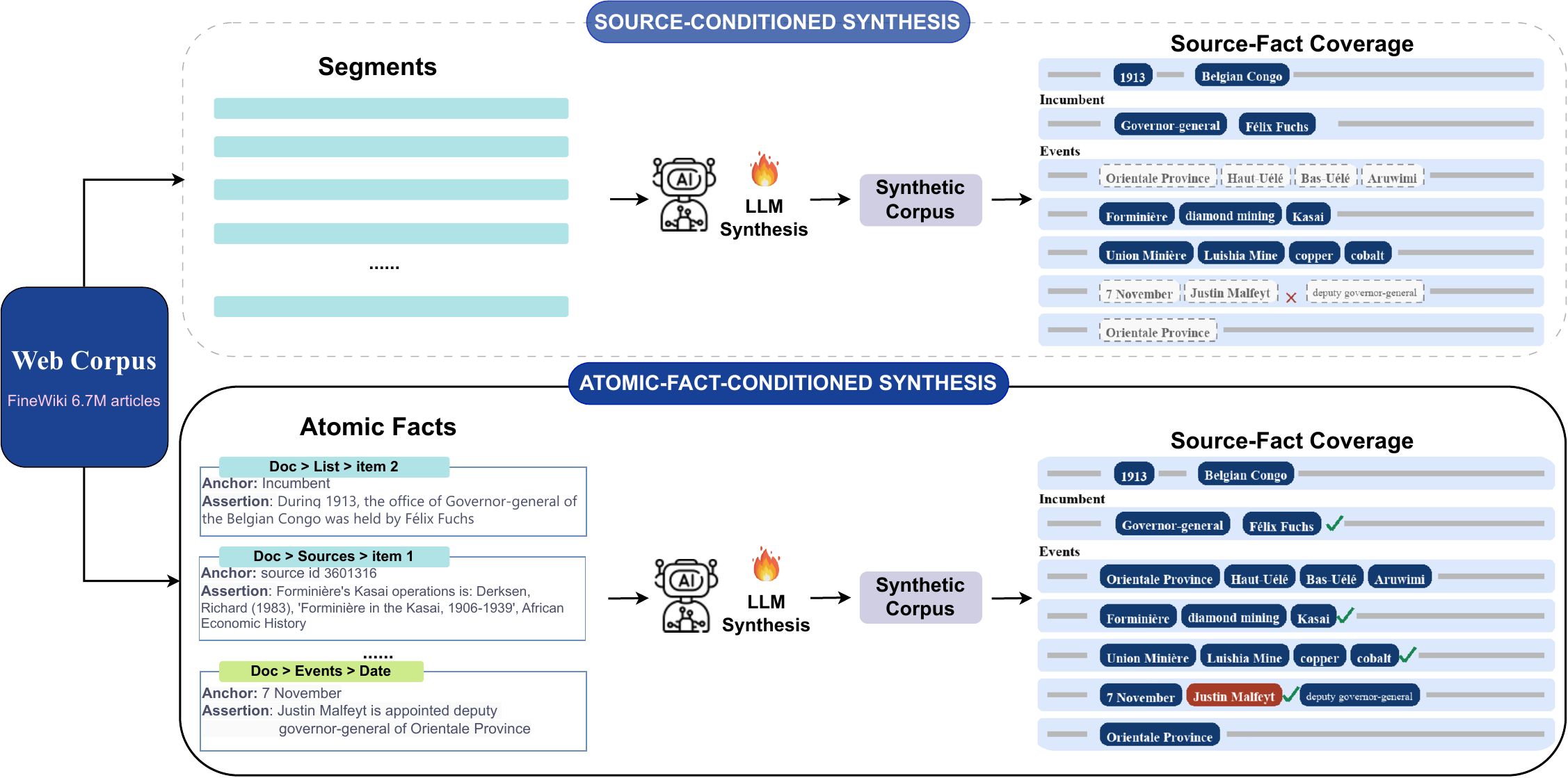}
\caption{\textbf{AFCS pipeline and source-fact coverage.} Source-conditioned synthesis (top) segments the web corpus and asks an LLM to synthesize from each segment, so the generated corpus tends to preserve salient facts while under-covering fine-grained facts in lists, tables, and secondary mentions. Atomic-fact-conditioned synthesis (bottom) first extracts structure-aware atomic facts from the same source corpus and then conditions generation on source-anchored assertion cards. Each card marks the local anchor (e.g., a document scope, list item, table row, or date cell), the natural-language assertion, and the highlighted value structure (e.g., a role binding, multi-element set, compound clause, or bibliographic record). These cards are generation targets rather than canonical entity--relation KG triples.}

\label{fig:afcs-pipeline}
\end{figure}

\section{Method: AFCS}
\label{sec:method}

AFCS is an atomic-fact-conditioned synthesis pipeline (Figure~\ref{fig:afcs-pipeline}). It extracts source-supported atomic facts, expands local context where needed, and generates training samples conditioned on extracted facts. This shifts fact selection upstream to extraction and verification, rather than leaving it entirely to the generator's preference for salient source content.

\subsection{Atomic Fact Extraction}
\label{sec:extraction}

Wikipedia pages have a natural document tree, including title, sections, paragraphs, lists, tables, and infoboxes. AFCS uses this structure during extraction. Tables, lists, numeric-heavy content, and prose are processed with separate prompt templates (Appendix~\ref{app:prompts}). Extraction statistics and author-checked precision/recall are reported in Appendix~\ref{app:extraction-stats}.

As defined in \S\ref{sec:atomic-facts}, an atomic fact is a source-supported assertion with a target value and local evidence span, not necessarily a canonical entity-centered record. AFCS represents dates, numerical attributes, aliases, list entries, table rows, and descriptive properties as typed atomic facts, without requiring a strict tuple schema. An AFCS unit is assertion-centric rather than entity-centric: it contains the statement to be realized, its target value, and the local evidence span used for verification and realization. Thus, entity attributes are not merely exposed as part of an entity profile; each attribute-value assertion can become an independent synthesis target. The same representation also covers non-canonical assertions such as evidential absence, conditional statements, causal explanations, reported speech, and full-clause values. When a fact depends on nearby context, AFCS expands the visible context around the focal span. When the fact spans related sections, those sections are opened while unrelated sections remain closed. The synthesis unit remains the extracted assertion or evidence span rather than the whole article.

\subsection{Source-Grounded Verification}
\label{sec:verification}

The extracted records are checked against their source spans before synthesis. This step filters out records whose values are judged unsupported by the verifier and preserves the source offset used for later context expansion. The verifier is not intended to produce a complete factual database. Its role is to reduce unsupported facts before they are amplified into multiple synthetic samples.

\subsection{Multi-Format Generation}
\label{sec:multiformat}

For each retained fact $f$ selected for synthesis and its local context $\mathrm{ctx}(f)$, AFCS generates complementary natural-language samples. All formats are driven by the extracted fact rather than by an unconstrained source-text prompt:
\begin{itemize}[nosep,leftmargin=*]
\item \textbf{Summary}: contextual summaries that preserve nearby factual relations and ordering.
\item \textbf{Rewrite}: paraphrases in alternative registers while preserving the same factual sequence.
\item \textbf{Continuation}: continuations from a local prefix that must include the target fact.
\item \textbf{QA}: answer-guided question generation. The value of the extracted fact is supplied as the target answer, and the generator writes natural questions whose answer is that value.
\end{itemize}

The sample-count and token-volume distribution across formats is reported in Appendix~\ref{app:sample-stats}.

\subsection{Corpus-Scale Construction}
\label{sec:corpus-scale}

We instantiate AFCS on approximately 6.7M FineWiki articles. Structure-aware extraction produces 312.7M atomic facts, with 34.9\% from prose, 29.2\% from tables or infoboxes, 19.6\% from numeric attributes, and 16.3\% from lists. The resulting AFCS corpus contains 142M natural-language synthesized samples across four formats. On a 500-article author-checked subset, extracted facts reach 94.1\% precision and 87.6\% recall; recall on list- and table-embedded facts is 82.4\%, compared with 48.7\% for a single-pass non-structure-aware extractor. These statistics characterize the scale and extraction quality of AFCS; comparative source-fact coverage is evaluated separately in Section~\ref{sec:coverage-results}.

\paragraph{Reproducibility details.} The AFCS extractor, source-grounded verifier, and synthetic-sample generator use the same instruction-tuned generator checkpoint, Qwen3-30B-A3B, served with vLLM. Extraction and verification use deterministic decoding (temperature 0), while multi-format generation uses temperature 0.7 with prompt-specific maximum output lengths. We filter records rejected by the source-grounded verifier and remove degenerate or condition-leaking generations. QA samples are deduplicated by exact question match, by fact id with at most two retained questions per fact, and by character-trigram overlap; multi-format outputs are deduplicated first by input span and then by output text. The final corpus contains 142M synthetic samples; the main 7B and 32B comparisons consume 20B packed training tokens from the corresponding synthetic corpus, and the mid-training integration uses 5B AFCS tokens. Data generation is a one-time offline inference cost separate from continued-pretraining compute.

\section{Experiments}
\label{sec:experiments}

\subsection{Experimental Setup}
\label{sec:setup}

\paragraph{Base model and training recipe.} All 7B methods continue training from OLMo-3-7B \citep{olmo2025olmo3} (stage 1, step 1.4M), with optimizer state restored from the base checkpoint. We use AdamW ($\beta_1{=}0.9,\beta_2{=}0.95$), peak learning rate $2.07\times10^{-5}$, linear decay to 0, batch size 2M tokens, and sequence length 8192. Before packing, training samples are globally shuffled and then concatenated into fixed-length windows. Runs use the actual synthesized corpora produced by each method, with generated-token counts reported in Table~\ref{tab:methods}; all main comparisons use a matched 20B-token continued-pretraining budget.

\paragraph{Compared methods.} All methods use the same FineWiki source corpus and the same base training recipe. Our WRAP baseline applies the four WRAP style prompts to each source passage or paragraph-level example \citep{maini2024rephrasing}. Our Nemotron-CC baseline reruns on FineWiki the five prompt types defined by Nemotron-CC \citep{su2025nemotron}. AFCS uses multi-format samples generated from extracted atomic facts. For the granularity ablations, chunk- and paragraph-level synthesis use the same generator with only the conditioning unit changed, while the entity-level baseline first extracts canonical subject--predicate--value triples and then conditions generation on both the extracted triple and its corresponding source context.

\begin{table}[t]
\centering\small
\caption{Compared methods. Gen. tok. counts generated tokens before packing; Train tok. is the matched continued-pretraining budget after packing, with corpora reshuffled/reused as needed. All runs use the same base model, source corpus, optimizer, and LR schedule.}
\label{tab:methods}
\setlength{\tabcolsep}{5pt}
\begin{tabular*}{\textwidth}{@{\extracolsep{\fill}}lllll}
\toprule
\textbf{Method} & \textbf{Data type} & \textbf{Gen. tok.} & \textbf{Train tok.} & \textbf{Mode} \\
\midrule
WRAP & 4-prompt mix & 16.6B & 20B & source-cond. \\
Nemotron-CC & 5-prompt mix & 18.7B & 20B & source-cond. \\
\textbf{AFCS} & multi-format & 19.3B & 20B & \textbf{atomic-fact-cond.} \\
\bottomrule
\end{tabular*}
\end{table}

\paragraph{Evaluation.} We evaluate with SimpleQA, SimpleQA Verified, PopQA, and HotpotQA using few-shot in-context learning and judge-based accuracy; the benchmark sources are cited in Section~\ref{sec:related}. Few-shot ICL is a standard way to evaluate pretrained base models without task-specific finetuning; here it probes factual acquisition and recoverability after continued pretraining. The specific prompting, inference, judging, sampling, and scoring settings are provided in Appendix~\ref{app:bench}.

\subsection{Overall Factual QA Performance}
\label{sec:main_results}

\begin{table}[!t]
\centering\small
\caption{Main factual benchmark results for synthetic-data methods at 7B and 32B scale. Tokens denotes continued-pretraining tokens; Base is evaluated before continued pretraining. Values are judge-based accuracies (\%).}
\label{tab:main}
\setlength{\tabcolsep}{3pt}
\begin{tabular*}{\textwidth}{@{\extracolsep{\fill}}lllcccc}
\toprule
\textbf{Model} & \textbf{Size} & \textbf{Tokens} & \textbf{SimpleQA} & \textbf{SimpleQA Verified} & \textbf{PopQA} & \textbf{HotpotQA} \\
\midrule
Base & 7B & -- & 3.63 & 3.95 & 23.06 & 27.78 \\
WRAP & 7B & 20B & 3.99 & 4.22 & 25.74 & 27.08 \\
Nemotron-CC & 7B & 20B & 3.99 & 4.16 & 26.42 & 27.40 \\
\textbf{AFCS} & \textbf{7B} & \textbf{20B} & \textbf{5.63} & \textbf{6.17} & \textbf{29.80} & \textbf{29.82} \\
\midrule
Base & 32B & -- & 5.94 & 5.78 & 29.17 & 34.50 \\
WRAP & 32B & 20B & 7.10 & 7.20 & 32.46 & 35.07 \\
Nemotron-CC & 32B & 20B & 6.82 & 7.24 & 33.01 & 34.60 \\
\textbf{AFCS} & \textbf{32B} & \textbf{20B} & \textbf{9.11} & \textbf{9.08} & \textbf{37.22} & \textbf{35.40} \\
\bottomrule
\end{tabular*}
\end{table}

\begin{figure}[!t]
\centering
\includegraphics[width=0.82\linewidth]{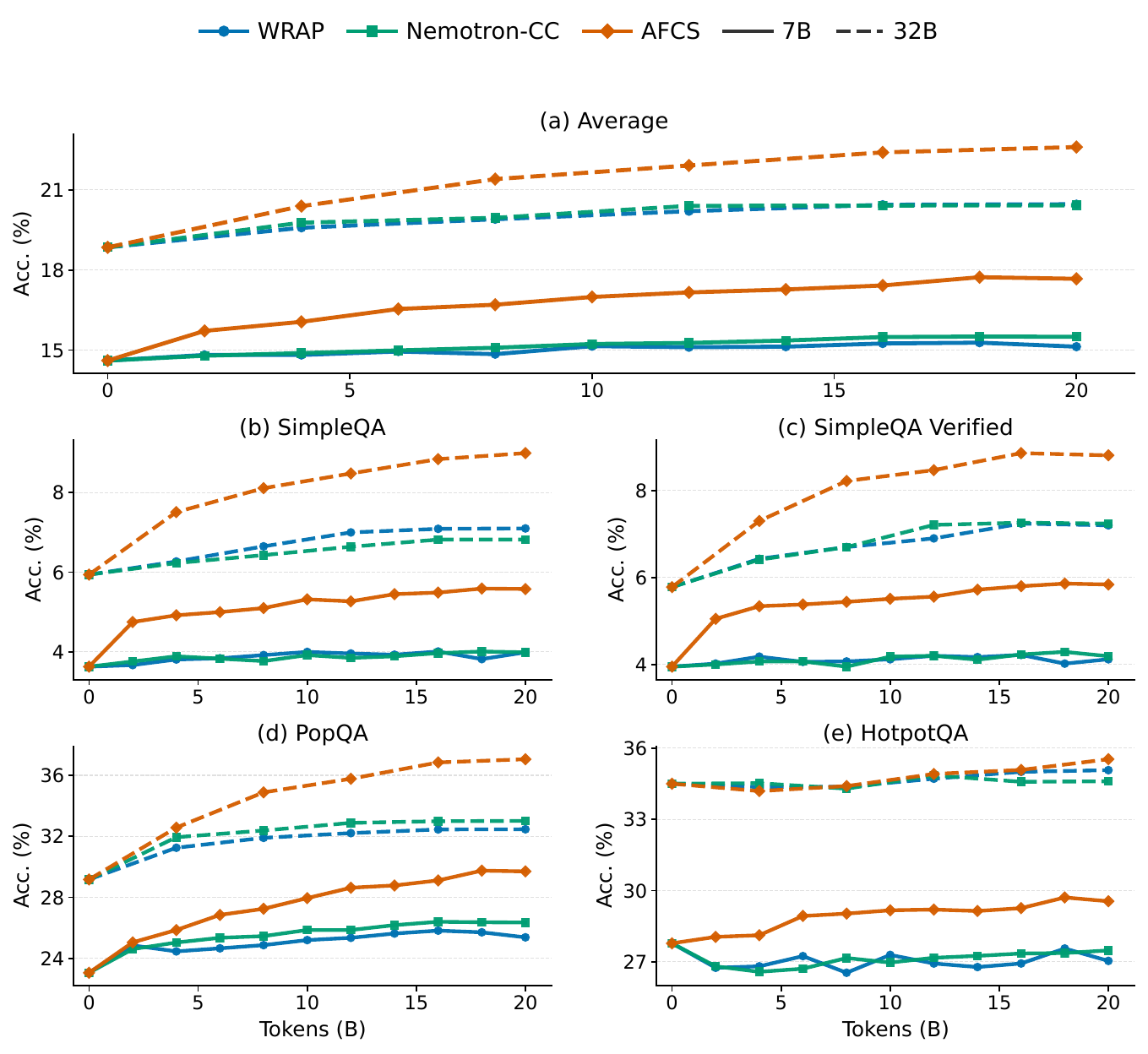}
\caption{Training dynamics across factual QA benchmarks. The x-axis is aligned by consumed training tokens: 1k steps correspond to 2B tokens for 7B models and 4B tokens for 32B models. Solid lines denote 7B models and dashed lines denote 32B models. AFCS maintains consistent gains over source-conditioned baselines across the average score and individual factual QA benchmarks.}
\label{fig:training-curves}
\end{figure}

Across 7B and 32B scales, AFCS obtains the best results among the synthetic-data methods on all four factual benchmarks (Table~\ref{tab:main}). At 7B, it improves over the base by $+2.00$ on SimpleQA, $+2.22$ on SimpleQA Verified, $+6.74$ on PopQA, and $+2.04$ on HotpotQA. At 32B, the corresponding gains over the base are $+3.17$, $+3.30$, $+8.05$, and $+0.90$. The largest gains appear on PopQA, consistent with improved factual recall in this setting. Figure~\ref{fig:training-curves} shows that the advantage persists throughout training when progress is aligned by consumed tokens. FineWiki Raw controls at both model scales are reported in Appendix~\ref{app:raw-baseline}.

\subsection{Explicit Source-Fact Coverage}
\label{sec:coverage-results}

We additionally measure whether sampled source-supported facts are explicitly realized in the synthetic outputs derived from the same document. Under this data-side diagnostic, AFCS reaches 85.1\% source-fact coverage, compared with 73.0\% for WRAP and 77.0\% for Nemotron-CC. Table~\ref{tab:coverage} reports the breakdowns by document fact density and by the fact types highlighted in the omission analysis; the evaluation protocol, aggregation rule, and natural prose-only subset are provided in Appendix~\ref{app:coverage-eval}. This diagnostic characterizes the synthetic corpus; it does not imply that every covered fact is acquired or retrieved by the trained model.

\begin{table}[t]
\centering\small
\caption{Source-fact coverage (\%) by document fact density and for the fact types highlighted in the omission analysis.}
\label{tab:coverage}
\setlength{\tabcolsep}{5pt}
\begin{tabular*}{\textwidth}{@{\extracolsep{\fill}}llrrrrrr}
\toprule
\textbf{Breakdown} & \textbf{Group} & \textbf{AFCS} & \textbf{WRAP} & \textbf{Nemo.} & \textbf{Chunk} & \textbf{Para.} & \textbf{Entity} \\
\midrule
\multirow{3}{*}{Facts/doc}
 & 1--19 & \textbf{90.3} & 82.2 & 83.5 & 80.4 & 77.3 & 70.3 \\
 & 20--99 & \textbf{88.9} & 82.6 & 84.6 & 78.9 & 73.3 & 65.2 \\
 & 100+ & \textbf{82.5} & 61.2 & 66.5 & 68.8 & 69.6 & 58.4 \\
\midrule
\multirow{3}{*}{Fact type}
 & List/table & \textbf{82.3} & 72.9 & 78.8 & 70.9 & 70.0 & 53.5 \\
 & Secondary entity & \textbf{83.6} & 73.4 & 79.9 & 70.5 & 69.6 & 52.0 \\
 & Fine-grained temporal & \textbf{80.8} & 72.5 & 78.1 & 69.8 & 69.5 & 53.4 \\
\bottomrule
\end{tabular*}
\end{table}

\subsection{Component Ablations}
\label{sec:ablation}

We ablate synthesis granularity and the mechanisms within AFCS on the 7B last checkpoint. The conditioning-unit block compares fact-level conditioning against coarser alternatives using the shared model and training/evaluation setup. In the entity variant, extraction is explicitly constrained to canonical subject--predicate--value triples; generation then receives the triple together with its source context, providing a matched-budget entity/triple-conditioned comparison with an alternative representation unit. We also adapt EntiGraph relation synthesis \citep{yang2024synthetic} and the fact-to-QA pipeline of \citet{mecklenburg2024injecting} under the same 20B-token CPT budget and evaluation setup. The mechanism block provides matched-budget contrasts for output format, per-fact amplification, and fact coverage.

\begin{table}[!t]
\centering\small
\caption{Granularity, adapted-baseline, and mechanism comparisons at 7B scale under the same 20B-token CPT budget. Values are percentages; SQA-V denotes SimpleQA Verified. Fact (AFCS) is the full multi-format mixture, and w/o verification removes source-grounded verification from it. AFCS-QA retains all facts with multiple QA realizations; 1 QA / fact reduces per-fact amplification; Half facts retains 50\% of the facts.}
\label{tab:ablation}
\setlength{\tabcolsep}{5pt}
\begin{tabular*}{\textwidth}{@{\extracolsep{\fill}}llcccc}
\toprule
\textbf{Block} & \textbf{Variant} & \textbf{SQA} & \textbf{SQA-V} & \textbf{PopQA} & \textbf{Hotpot} \\
\midrule
Reference & Base & 3.63 & 3.95 & 23.06 & 27.78 \\
\midrule
\multirow{4}{*}{Unit}
 & \textbf{Fact (AFCS)} & \textbf{5.63} & \textbf{6.17} & \textbf{29.80} & \textbf{29.82} \\
 & Chunk & 4.21 & 4.80 & 27.43 & 28.94 \\
 & Paragraph & 4.52 & 5.10 & 26.35 & 27.61 \\
 & Entity/triple & 3.85 & 4.30 & 25.10 & 27.20 \\
\midrule
\multirow{2}{*}{Baseline}
 & EntiGraph & 3.91 & 4.56 & 23.53 & 26.56 \\
 & Fact-QA & 4.82 & 5.31 & 25.34 & 28.98 \\
\midrule
\multirow{4}{*}{Mechanism}
 & w/o verification & 5.49 & 6.00 & 29.77 & 29.60 \\
 & AFCS-QA & 5.32 & 6.00 & 27.03 & 29.14 \\
 & 1 QA / fact & 4.90 & 5.55 & 25.72 & 28.22 \\
 & Half facts & 4.77 & 5.52 & 24.84 & 27.60 \\
\bottomrule
\end{tabular*}
\end{table}

Among the matched comparisons, fact-level AFCS is strongest, improving over chunk-level synthesis by $+1.42/+1.37/+2.37/+0.88$ on SimpleQA, SimpleQA Verified, PopQA, and HotpotQA. AFCS also remains above the adapted EntiGraph and Fact-QA baselines. The full multi-format mixture outperforms AFCS-QA in this setup. AFCS-QA in turn improves over 1 QA / fact and Half facts, consistent with benefits from repeated realizations per fact and broader fact coverage, respectively.

\paragraph{Matched-budget mechanism contrasts.}
The mechanism rows in Table~\ref{tab:ablation} use the same CPT token budget. The 1 QA / fact variant retains all facts but limits each fact to one QA realization, while Half facts retains 50\% of the facts and assigns the retained facts a correspondingly larger share of QA samples. These comparisons vary per-fact amplification and source-fact breadth while keeping the overall training budget aligned.

Removing source-grounded verification changes AFCS from 5.63\slash 6.17\slash 29.80\slash 29.82 to 5.49\slash 6.00\slash 29.77\slash 29.60 on SimpleQA, SimpleQA Verified, PopQA, and HotpotQA, respectively. These differences are smaller than the granularity contrasts in Table~\ref{tab:ablation}; verification provides modest gains in this setup.

On SimpleQA Verified, AFCS is consistently strongest for pass@$k$ at both 7B and 32B scale (Table~\ref{tab:sqa-verified-passk}). Here, pass@$k$ is the proportion of questions for which at least one of $k$ generated answers is correct. Pass@1 measures deterministic single-answer accuracy, while larger $k$ values measure whether a correct answer is recoverable across multiple sampled generations; these results provide model-side evidence complementary to the data-side coverage diagnostic.

\begin{table}[t]
\centering
\begin{minipage}[b]{0.475\linewidth}
\centering\small
\caption{SimpleQA Verified pass@$k$ (\%).}
\label{tab:sqa-verified-passk}
\footnotesize
\setlength{\tabcolsep}{3pt}
\begin{tabular}{llccccc}
\toprule
\textbf{Scale} & \textbf{Model} & \textbf{@1} & \textbf{@8} & \textbf{@16} & \textbf{@32} & \textbf{@64} \\
\midrule
\multirow{4}{*}{7B}
 & Base & 3.95 & 13.52 & 18.28 & 23.75 & 29.60 \\
 & WRAP & 4.22 & 15.73 & 21.78 & 28.26 & 34.50 \\
 & Nemotron & 4.16 & 15.50 & 21.51 & 27.99 & 34.40 \\
 & \textbf{AFCS} & \textbf{6.17} & \textbf{18.96} & \textbf{24.94} & \textbf{31.28} & \textbf{37.50} \\
\midrule
\multirow{4}{*}{32B}
 & Base & 5.78 & 18.05 & 23.57 & 29.47 & 35.30 \\
 & WRAP & 7.20 & 21.39 & 27.50 & 33.68 & 39.60 \\
 & Nemotron & 7.24 & 21.53 & 27.89 & 34.57 & 40.90 \\
 & \textbf{AFCS} & \textbf{9.08} & \textbf{24.16} & \textbf{30.91} & \textbf{37.95} & \textbf{44.30} \\
\bottomrule
\end{tabular}
\end{minipage}\hfill
\begin{minipage}[b]{0.505\linewidth}
\centering\small
\caption{PopQA judge-based accuracy (\%) by monthly page-view bucket.}
\label{tab:popqa}
\footnotesize
\setlength{\tabcolsep}{3pt}
\begin{tabular}{llrrrrr}
\toprule
\textbf{Scale} & \textbf{Views} & \textbf{N} & \textbf{Base} & \textbf{WRAP} & \textbf{Nemo.} & \textbf{AFCS} \\
\midrule
\multirow{4}{*}{7B}
 & 0--100 & 1{,}399 & 25.1 & 32.5 & 32.5 & \textbf{38.2} \\
 & 100--1K & 5{,}791 & 11.6 & 14.7 & 14.7 & \textbf{20.6} \\
 & 1K--10K & 4{,}259 & 19.6 & 22.0 & 23.7 & \textbf{26.1} \\
 & 10K+ & 2{,}807 & 50.6 & 48.7 & \textbf{51.1} & 49.9 \\
\midrule
\multirow{4}{*}{32B}
 & 0--100 & 1{,}399 & 28.6 & 32.6 & 34.2 & \textbf{40.3} \\
 & 100--1K & 5{,}791 & 16.0 & 19.0 & 19.6 & \textbf{26.6} \\
 & 1K--10K & 4{,}259 & 29.0 & 33.1 & 33.4 & \textbf{36.8} \\
 & 10K+ & 2{,}807 & 56.7 & 59.1 & \textbf{59.3} & 58.1 \\
\bottomrule
\end{tabular}
\end{minipage}
\end{table}

\subsection{Popularity-Stratified PopQA Analysis}
\label{sec:popqa}

We stratify PopQA by entity monthly page views and apply the same judge-based scoring protocol used for the aggregate result (Table~\ref{tab:popqa}).

AFCS gains are largest on rare and middle-tail entities: over WRAP, it improves by $+5.7/+5.9$ points in the two lowest buckets at 7B and by $+7.7/+7.6$ points at 32B. In the 10K+ bucket, methods are close, suggesting that prompt diversity remains competitive for popular entities.

\subsection{Scaling to 32B}
\label{sec:scale32b}

We rerun the main synthesis variants on a 32B base with the same FineWiki source corpus and matched training recipe. AFCS remains strongest among the synthetic-data methods on all four metrics, with the largest 32B gain on PopQA ($+8.05$ over Base, $+4.76$ over WRAP, and $+4.21$ over Nemotron-CC).

\paragraph{Setup.} The 32B continued-pretraining runs use the same FineWiki source corpus and the same 20B-token budget as the 7B runs (\S\ref{sec:setup}); peak LR is rescaled to $6.9\%$ of the 32B stage-1 peak following the base recipe, batch size 4M tokens, sequence length 8192, AdamW ($\beta_1{=}0.9,\beta_2{=}0.95$), linear decay to 0. We compare AFCS, WRAP, and Nemotron-CC.

\subsection{Integration with OLMo-3 Mid-Training Data}
\label{sec:midtrain}

We test whether AFCS can be integrated into a 100B-token mid-training mixture by replacing 5B general tokens with AFCS data and evaluating both factual QA and the 12-task general suite (Table~\ref{tab:midtrain}).

\begin{table}[!t]
\centering\small
\caption{Integrating AFCS into OLMo-3-7B 100B-token mid-training. Values are percentages; SQA-V denotes SimpleQA Verified.}
\label{tab:midtrain}
\setlength{\tabcolsep}{5pt}
\begin{tabular*}{\textwidth}{@{\extracolsep{\fill}}llccccc}
\toprule
\textbf{Model} & \textbf{Data} & \textbf{SQA} & \textbf{SQA-V} & \textbf{PopQA} & \textbf{Hotpot} & \textbf{12-task} \\
\midrule
Base & -- & 3.63 & 3.95 & 23.06 & 27.78 & 57.79 \\
Midtrain & 100B gen. & 3.68 & 4.02 & 23.18 & 28.07 & 68.24 \\
\textbf{Mid.+AFCS} & \textbf{95B gen.+5B AFCS} & \textbf{4.36} & \textbf{4.69} & \textbf{26.06} & \textbf{32.14} & \textbf{69.33} \\
\bottomrule
\end{tabular*}
\end{table}

Replacing 5B general tokens with AFCS improves SimpleQA, SimpleQA Verified, PopQA, HotpotQA, and the 12-task average by $+0.68/+0.67/+2.88/+4.07/+1.09$, respectively.

Per-benchmark results in Appendix~\ref{app:midtrain-breakdown}, Table~\ref{tab:midtrain-perbenchmark}, show gains on SQuAD v2, WikiTableQuestions, TriviaQA, PIQA, and HellaSwag; five of the twelve tasks decrease by 0.33--2.80 points. Overall, the reported 12-task average increases by $+1.09$.

\section{Discussion}
\label{sec:analysis}

\subsection{Why Synthesis Granularity Matters}
\label{sec:granularity}

The granularity rows in Table~\ref{tab:ablation} show that fact-level conditioning is more effective than chunk-level, paragraph-level, or entity-level synthesis. Coarser units preserve local context, but they again leave the generator to select which fine-grained facts to express. Entity-level synthesis narrows the topic but still does not specify the relation or attribute that must be covered. Atomic facts provide a smaller and more explicit conditioning unit, making the target value directly visible while retaining enough source context for fluent realization.

\subsection{Source-Conditioned Baselines}
\label{sec:baseline_discussion}

The comparison with WRAP and Nemotron-CC indicates that prompt diversity alone does not fully address fact selection. WRAP targets surface-form variation but yields smaller PopQA gains in our setup. Nemotron-CC performs well on average and on some popularity buckets, yet it still conditions generation on document segments rather than extracted facts. AFCS differs by making the target fact explicit before generation. The largest gains appear in the rare and middle-frequency PopQA buckets.

\subsection{Coverage, Acquisition, and Scope}
\label{sec:coverage-scope}

Explicit source-fact coverage is a property of the synthesized corpus, whereas factual QA and pass@$k$ measure model-side acquisition and recoverability after continued pretraining. The higher coverage and pass@$k$ observed for AFCS provide consistent but indirect evidence across these two levels; they are not a one-to-one measurement that each covered fact is stored and retrieved. Accordingly, we interpret the corpus-coverage and model-performance measures as complementary rather than as establishing a causal link.

\section{Conclusion}

We studied a coverage failure mode in knowledge-corpus synthesis: when generation is conditioned on a source passage or document segment, fine-grained facts present in the source may be omitted from the synthetic corpus. AFCS mitigates this issue by extracting atomic facts and using retained facts as generation targets. On FineWiki, AFCS improves factual QA performance at 7B and is strongest on middle-tail PopQA entities. Matched-budget granularity ablations show that fact-level conditioning outperforms coarser units, while targeted mechanism contrasts yield results consistent with benefits from broader fact coverage, repeated realizations per fact, and multi-format generation. The 32B results show that AFCS remains strongest among the synthetic-data methods across SimpleQA, SimpleQA Verified, PopQA, and HotpotQA. In a 100B-token OLMo-3 mid-training integration, AFCS improves factual QA benchmarks and raises the 12-task general benchmark average to 69.33. Future work includes stronger extraction verification, multilingual extension, and combining atomic-fact-conditioned synthesis with corpus filtering and retrieval.

\section*{Limitations}

\begin{itemize}[nosep,leftmargin=*]
\item \textbf{Extractor dependence.} AFCS depends on an LLM extractor and verifier. We report precision and recall on an author-checked subset, but residual extraction errors can propagate to synthetic samples.
\item \textbf{Coverage versus learning.} Explicit presence in the synthetic corpus does not guarantee that the trained model stores or retrieves a fact.
\item \textbf{Scale.} We evaluate 7B and 32B models. Whether the same trends persist at larger model scales remains open.
\item \textbf{Domain.} All experiments use English Wikipedia. Generalization to other languages and domains such as scientific literature, legal text, or code is untested.
\item \textbf{Cost.} AFCS requires structure-aware extraction, verification, and multi-format generation. This is a one-time data-construction cost and incurs additional offline inference at the multi-million-article scale.
\item \textbf{Retrieval scope.} AFCS changes pretraining data to improve parametric factual knowledge and does not compare against retrieval-augmented systems. The two approaches address different deployment trade-offs and may be combined in future work.
\end{itemize}

\section*{Ethics Statement}

Our training corpus is English Wikipedia, a publicly licensed knowledge resource. Atomic facts and synthesized samples inherit Wikipedia's existing biases toward over-represented regions, languages, and demographic groups. AFCS may amplify facts present in the source corpus, including those biases. We do not perform additional debiasing or fact-checking beyond Wikipedia's editorial process and our source-grounded verification step. Synthesized samples may still contain unsupported surface phrasing and should be used for pretraining research rather than as a standalone factual reference.

\FloatBarrier
\begingroup\small
\bibliography{refs}
\endgroup

\clearpage
\appendix
\suppressfloats[t]

\section{Atomic-Fact Extraction Statistics}
\label{app:extraction-stats}

\paragraph{Pipeline scale.} We process approximately 6.7M FineWiki articles. After section segmentation, structure-aware extraction yields the following per-article fact counts:

\begin{table}[H]
\centering\small
\caption{Atomic-fact extraction statistics over the FineWiki training corpus.}
\label{tab:extraction-stats}
\begin{tabular}{lr}
\toprule
\textbf{Statistic} & \textbf{Value} \\
\midrule
Articles processed & $\sim$6.7M \\
Sections processed & 78.4M \\
Total atomic facts extracted & 312.7M \\
Mean facts / article & 46.7 \\
Median facts / article & 18 \\
P5 / P95 facts / article & 3 / 92 \\
\midrule
\% facts from prose & 34.9 \\
\% facts from tables / infoboxes & 29.2 \\
\% facts from numeric attributes & 19.6 \\
\% facts from lists & 16.3 \\
\bottomrule
\end{tabular}
\end{table}

\paragraph{Extraction quality.} On a 500-article author-checked subset, we report:
\begin{itemize}[nosep,leftmargin=*]
\item Precision (extracted facts that are correct \& verifiable in source): $94.1\%$
\item Recall (verifiable atomic facts in source that are extracted): $87.6\%$
\item Recall on list-/table-embedded facts specifically: $82.4\%$ (vs.\ $48.7\%$ for a single-pass non-structure-aware extractor)
\end{itemize}

\section{Examples of Non-Canonical Atomic Assertions}
\label{app:noncanonical}

Table~\ref{tab:noncanonical} illustrates why the subject--predicate--value notation used by AFCS should not be read as a canonical KG triple schema. These examples can often be encoded in richer graph form using reification, qualifiers, or custom schemas, but they are not naturally represented as a single entity--relation edge. AFCS uses them directly as source-local generation targets, preserving the local evidence span and target value without requiring corpus-level schema canonicalization.

\begin{table}[t]
\centering\small
\caption{Examples of AFCS atomic assertions that are source-local generation targets rather than canonical KG triples. The notation is compact, but the predicate and value may include negation, attribution, causality, conditionals, or full clauses.}
\label{tab:noncanonical}
\setlength{\tabcolsep}{3pt}
\begin{tabular*}{\textwidth}{@{\extracolsep{\fill}}>{\raggedright\arraybackslash}p{0.22\textwidth}>{\raggedright\arraybackslash}p{0.43\textwidth}>{\raggedright\arraybackslash}p{0.26\textwidth}}
\toprule
\textbf{Local anchor} & \textbf{Extracted source-supported assertion} & \textbf{Why it is non-canonical} \\
\midrule
Giovanni Boccaccio & observed that no history of Hippo's ancestry or place of birth survived, other than from ``books of the ancients'' & evidential absence and reported observation, not a positive entity attribute \\
Edzard Ernst & wrote that zero balancing did not appear to have evidence published for its efficacy & negation plus attribution to a source \\
Ng Sui & trained in mainland China because equipment in Hong Kong was not up to standard and training time was insufficient & causal explanation with an embedded comparison \\
Nuclear industry & where wastewater discharges to rivers with drinking-water abstraction, harmful or long-half-life radioisotopes such as tritium form part of routine monitoring & conditional process statement with a full-clause value \\
Segre theorem & states that for any point in a convex hull that is not a vertex, at least four curve points have osculating planes passing through it & mathematical assertion whose anchor is a theorem, not a named-entity relation \\
List of power pop artists and songs & includes power pop songs by artists whose main music genre is not power pop & list-scope assertion anchored by a document section \\
\bottomrule
\end{tabular*}
\end{table}

\section{Synthesized Sample Statistics}
\label{app:sample-stats}

\paragraph{Composition.} The 142M natural-language AFCS training samples decompose as shown in Table~\ref{tab:sample-comp}.

\paragraph{Length distributions (tokens per sample).} Table~\ref{tab:sample-length} reports per-sample token-length percentiles by format.

\begin{table}[t]
\centering
\begin{minipage}[t]{0.48\linewidth}
\centering\small
\caption{Sample-count and token-volume distribution by natural-language format.}
\label{tab:sample-comp}
\footnotesize
\setlength{\tabcolsep}{4pt}
\begin{tabular}{lrrr}
\toprule
\textbf{Format} & \textbf{Samples} & \textbf{\%} & \textbf{Tokens \%} \\
\midrule
Summary & 84M & 59.2 & 52.0 \\
QA (answer-guided) & 33M & 23.2 & 11.7 \\
Continuation & 17M & 12.0 & 21.4 \\
Rewrite & 8M & 5.6 & 15.0 \\
\midrule
Total & 142M & 100 & 100 \\
\bottomrule
\end{tabular}
\end{minipage}\hfill
\begin{minipage}[t]{0.48\linewidth}
\centering\small
\caption{Per-sample token-length percentiles by format.}
\label{tab:sample-length}
\footnotesize
\setlength{\tabcolsep}{4pt}
\begin{tabular}{lrrrrr}
\toprule
\textbf{Format} & \textbf{P5} & \textbf{P25} & \textbf{med} & \textbf{P75} & \textbf{P95} \\
\midrule
Summary & 60 & 92 & 128 & 184 & 312 \\
QA & 14 & 22 & 31 & 44 & 78 \\
Continuation & 84 & 156 & 248 & 392 & 712 \\
Rewrite & 96 & 184 & 296 & 488 & 944 \\
\bottomrule
\end{tabular}
\end{minipage}
\end{table}

\paragraph{Source-article length.} Source articles have median 612 tokens, P5/P95 of 84 / 4{,}812 tokens. Article length correlates only weakly with per-article fact count ($r=0.41$) because fact density varies sharply between prose-heavy biographies and table-heavy list articles.

\section{Extraction, Verification, and Synthesis Prompt Templates}
\label{app:prompts}

We list the core prompt templates used for atomic-fact extraction, verification, and synthesis. Extraction templates take \texttt{\{title\}} and \texttt{\{document\_chunk\}} as inputs; subsequent stages additionally receive structured \texttt{\{atomic\_facts\}}.

\paragraph{Atomic-fact extraction (structure-aware).}
AFCS uses separate table, list, numeric, and prose extractor prompts. Braces in the templates denote runtime substitutions; repetitive in-prompt examples are omitted for space.

\paragraph{Generic no-structure mode.}
The \texttt{no\_struct} ablation replaces the four structure-specific prompts with a single generic extractor. It is distinct from the natural prose-only document subset reported in Table~\ref{tab:coverage-prose}.

\paragraph{AFCS answer-guided QA synthesis.}
The QA instruction is adapted from the QA-style prompt of WRAP \citep{maini2024rephrasing}, with the additional constraint that the answer is fixed to the value of an extracted fact rather than chosen by the generator.

\paragraph{Summary, Rewrite, and Continuation.}
The concrete multi-format prompt is used for the AFCS non-QA formats.

\paragraph{Granularity baselines.}
Chunk- and paragraph-level baselines call the same generator templates with the source chunk or paragraph as the conditioning input. The entity-level baseline uses a separate extractor prompt that explicitly asks for canonical subject--predicate--value triples only; its generator receives both the extracted triple and the corresponding local source context. This design provides an entity/triple-conditioned comparison, while AFCS conditions on source-anchored fact records that may include table rows, list items, dates, quantities, or full-clause values.

\paragraph{Fact-QA baseline construction.}
We adapt the two-stage passage-to-fact-to-QA structure of \citet{mecklenburg2024injecting}, including up to ten QA pairs per fact, optional \texttt{SKIP} for facts that are too broad or insufficiently specific, top-$p$ 0.95, and exact-match question deduplication. In our implementation, Qwen3-30B-A3B performs both stages with temperature 0.7 and a maximum of 1,500 output tokens per stage; exact duplicate questions are removed during packing. The prompt templates are reproduced below.

\begin{promptbox}{Table extractor}
Extract structured facts from this document. Output a JSON array only. The document is about: \{title\}. Each fact is one object: \{``subj'':``...'', ``pred'':``...'', ``val'':``...'', ``source\_span'':``...''\}. \par

\textbf{\sffamily Rules} \par
- Process every data row, including the final rows of long tables. For scorecard or statistics tables, extract every meaningful numeric column for every row. \par
- For tables with three or more columns, use multi-direction extraction: rotate which meaningful column is the answer value and place the remaining columns in the subject or predicate as context. Always retain the most natural direction; add reverse directions for entity names, people, places, awards, teams, years, and dates, but not for purely numeric attributes whose reverse direction is unnatural. \par
- For cast or crew entries, extract both actor-to-character and character-to-actor directions. \par
- Also capture dense structured records expressed without explicit table markup, including entity attributes, biographical details, events, and discography items. \par
- Preserve the exact supporting text in \texttt{source\_span}; keep \texttt{val} concise. Return at most 500 facts. Start with [ and end with ]; output no other text. \par

\textless document\textgreater\{document\_chunk\}\textless/document\textgreater
\end{promptbox}

\begin{promptbox}{List extractor}
Extract list, award, and rank items from the document. Output a JSON array only. The document is about: \{title\}. Each entry is one object: \{``subj'':``...'', ``pred'':``qualifier'', ``val'':``entry name'', ``source\_span'':``...''\}. \par

\textbf{\sffamily Rules} \par
- Set \texttt{subj} to the entity described by the list, not to a generic category label such as ``Awards'' or ``Works.'' Produce one object per entry and per applicable direction; never merge entries. \par
- For award/year lists, extract applicable recipient, year, and award-name directions so that each meaningful field can serve as the answer. Process every year, including early historical entries. \par
- For cast or crew lists, extract every member in both actor-to-character and character-to-actor directions. \par
- In Recognition, Awards, and Honors sections, extract all year--prize pairs and the corresponding recipient direction. Skip section headers and vague entries. \par
- Preserve the exact supporting text in \texttt{source\_span}. Return at most 500 facts. Start with [ and end with ]; output no other text. \par

\textless document\textgreater\{document\_chunk\}\textless/document\textgreater
\end{promptbox}

\begin{promptbox}{Numeric extractor}
Extract all dates, counts, measurements, rankings, codes, and product specifications from the document. The document is about: \{title\}. For each item, output \{``subj'':``...'', ``pred'':``...'', ``val'':``...'', ``source\_span'':``...''\}. \par

\textbf{\sffamily Rules} \par
- Use the specific entity described by the number, date, or code as \texttt{subj}. Prefer exact, question-answerable values and skip approximate or vague numbers. \par
- Extract years from recognition, award, and honor lists even when they appear in list format. \par
- For month--year and full-date mentions, emit both the full-granularity value and the year; for year-only mentions, emit one fact. Preserve exact counts and measurements without embedding explanatory phrases in \texttt{val}. \par
- Include secondary-event dates such as publication, appointment, certification, release, performance, competition, interview, statement, and speech dates. \par
- Include product codes, model identifiers, standards, serial or edition numbers, and exact small counts. Output only a valid JSON array. \par

\textless document\textgreater\{document\_chunk\}\textless/document\textgreater
\end{promptbox}

\begin{promptbox}{Prose extractor}
Extract (subject, predicate, value) facts from the document prose. The document is about: \{title\}. Focus on entity relationships, biographical details, events, and attributions. For each fact, output \texttt{subj}, \texttt{pred}, \texttt{val} (the answer), and \texttt{source\_span} (exact supporting text). Output only a JSON array and aim for completeness. \par

\textbf{\sffamily Rules} \par
- Use a specific named entity as \texttt{subj}; never use vague references such as ``the author'' or ``the subject.'' Prefer facts that support specific, unambiguous questions rather than general descriptions. \par
- Pay particular attention to appointment or nomination, competition or defeat, creation or authorship, replacement or succession, marriage or partnership, aliases, membership, event locations, requesters or initiators, education, and dated interviews or statements. \par
- For role portrayals, extract both actor-to-character and character-to-actor directions. For passive constructions such as ``X was appointed by Y,'' set \texttt{subj}=X, \texttt{pred}=``appointed by,'' and \texttt{val}=Y. \par

\textless document\textgreater\{document\_chunk\}\textless/document\textgreater
\end{promptbox}

\begin{promptbox}{Generic no-structure mode}
Extract all factual information from the document as (subject, predicate, value) facts. The document is about: \{title\}. Be comprehensive: include entities, dates, numbers, list items, and table values. For each fact, output \texttt{subj}, \texttt{pred}, \texttt{val}, and the exact \texttt{source\_span}. Use a specific named entity as the subject and focus on facts that can form specific, unambiguous questions. Output only a JSON array. \par

\textless document\textgreater\{document\_chunk\}\textless/document\textgreater
\end{promptbox}

\begin{promptbox}{Source-grounded verifier}
You will receive a source context and an extracted fact record: (anchor, predicate\_or\_attribute, value, span). Decide whether the value is directly supported by the source context. \par

\textbf{\sffamily Requirements} \par
- Use only the provided context. \par
- If supported, output \{"verdict":"SUPPORTED", "normalized\_value":"..."\}. \par
- If unsupported, output \{"verdict":"UNSUPPORTED", "reason":"..."\}. \par
- If the context is insufficient, output \{"verdict":"NOT\_FOUND"\}. \par

\textbf{\sffamily Output} \par
Return JSON only.
\end{promptbox}

\begin{promptbox}{AFCS answer-guided QA synthesis}
For EACH fact below, generate 1--2 question--answer pairs. The answer is fixed to the value of the extracted fact. Use the local source context as primary evidence, including the source span and any section-level breadcrumb needed to make the question specific. \par

\textbf{Rules} \par
1. The question must be answerable from the local context or the document. \par
2. Write natural questions; avoid overly formal forms such as ``Presupposing that ...''. \par
3. Do not create tautological QA where the question merely restates the answer. \par
4. Match question type to answer type: person names use Who; dates/years use When or What year; numbers use How many or How much; places use Where. \par
5. For list or table facts, include the year, row, position, institution, award, or other qualifier needed to identify the exact row/item. \par
6. Preserve temporal granularity: if the answer is a full date, ask for the full date; if it is month plus year, ask for month and year; if it is a year, include enough event context to avoid ambiguity. \par
7. If several facts in the batch have similar predicates but different answers, make each question discriminative by adding the subject name or another distinguishing qualifier. \par
8. Do not include the answer verbatim inside the question. \par

\textbf{Output} \par
Return JSON only: [\{``fact\_id'':``...'', ``question'':``...'', ``answer'':``...''\}].
\end{promptbox}

\begin{promptbox}{AFCS multi-format synthesis}
For each fact below, generate ONE training data sample in the best non-QA format. Use the document context and the fact fields \texttt{subj}, \texttt{pred}, \texttt{val}, \texttt{fact\_type}, and \texttt{local\_context}. \par

\textbf{Available formats} \par
1. \texttt{summary}: \{``fact\_id'':``...'', ``output\_type'':``summary'', ``input\_text'':``...'', ``summary'':``...''\}. \par
2. \texttt{rewrite}: \{``fact\_id'':``...'', ``output\_type'':``rewrite'', ``original'':``...'', ``rewritten'':``...'', ``style'':``simplified|formal|paraphrase''\}. \par
3. \texttt{continuation}: \{``fact\_id'':``...'', ``output\_type'':``continuation'', ``prefix'':``...'', ``continuation'':``...''\}. \par

\textbf{Output} \par
Return JSON only. Each output must express the target fact and remain supported by the provided local context.
\end{promptbox}

\begin{promptbox}{Fact-QA stage 1: passage-to-fact}
Please break down the following snippet from the Wikipedia article titled ``[THEME]'' into atomic facts. \par

Goal 1: The atomic facts should be as simple as possible, if it's compound sentence, break down one more time. \par
Goal 2: For clarity, avoid using pronouns like ``it'', ``he'', ``she'', ``this'', ``that'' etc., and instead use the full names or titles. \par

[PASSAGE] \par

1.
\end{promptbox}

\begin{promptbox}{Fact-QA stage 2: fact-to-QA}
Write 10 pairs of questions and answers probing the facts and statistics of a given sentence about [THEME]. \par
Consider first generating questions and answers that are very relevant and explicit to the fact, then paraphrase those questions and answers to reach the desired 10 Q\&A pairs. \par
If the fact is too broad or not specific enough to [THEME], you may reply with \texttt{SKIP} and be done. \par

EXAMPLE: \par
FACT: 14 million viewers tuned in to the opening game of the series. \par
1. Q: How many viewers watched the first game? A: 14 million people watched the first game of the series. \par

EXAMPLE: \par
FACT: The rose is red. \par
SKIP \par

FACT: [FACT] \par
1.
\end{promptbox}

\section{Source-Fact Coverage Evaluation}
\label{app:coverage-eval}

\paragraph{Protocol and sampling.} We use \texttt{gpt-oss-120b} \citep{openai2025gptoss} as the coverage judge. For each evaluated source-supported fact, the judge receives the synthetic data derived from the corresponding source document and determines whether the fact is explicitly covered. The diagnostic is run on 600K data instances; each instance pairs a reference fact with the method-specific synthetic outputs derived from the same source document. Each reference fact is represented by its subject, relation, value, source sentence, and section path.

\paragraph{Judge and aggregation.} A fact is marked as covered only when the exact claim, including its value, precision, and attribution, is recoverable from the synthetic text. Coverage is the proportion of evaluated facts judged covered. An author review of 100 judgments gives 96\% agreement with the automatic judge. This metric evaluates explicit coverage in the synthetic corpus rather than model acquisition.

\paragraph{Natural prose-only subset.} We identify 3,102,410 documents indexed by this diagnostic as natural prose-only sources and recompute coverage within this subset. This is a subset analysis rather than a rerun that removes structure from the same documents. AFCS remains above WRAP and Nemotron-CC within the prose-only subset, including the fact-dense group.

\begin{table}[H]
\centering\small
\caption{Source-fact coverage (\%) on the natural prose-only subset. The all-document AFCS column is included as a reference, not as a structure-removal intervention.}
\label{tab:coverage-prose}
\setlength{\tabcolsep}{5pt}
\begin{tabular*}{\textwidth}{@{\extracolsep{\fill}}lrrrr}
\toprule
\textbf{Facts/doc} & \textbf{AFCS (prose)} & \textbf{AFCS (all)} & \textbf{WRAP (prose)} & \textbf{Nemo. (prose)} \\
\midrule
1--19 & 89.1 & \textbf{90.3} & 82.2 & 83.5 \\
20--99 & 88.3 & \textbf{88.9} & 82.6 & 84.6 \\
100+ & 79.6 & \textbf{82.5} & 61.2 & 66.5 \\
\bottomrule
\end{tabular*}
\end{table}

\paragraph{Coverage-judge prompt.} The system and user messages are reproduced verbatim below; \texttt{\%\%CTX\%\%} and \texttt{\%\%FACTS\%\%} are the runtime placeholders for the document-specific synthetic text and numbered fact blocks.

\begin{promptbox}{Coverage-judge prompt}
\textbf{System:} You are a precise fact-checking assistant. \par

\textbf{User:} You are a STRICT fact-checking judge. You are given SYNTHETIC TEXT rewritten/derived from a Wikipedia article, and a NUMBERED LIST of atomic facts drawn from that same article. \par

Each fact is shown as subject / relation / value, plus the original source sentence it came from (and its section). Treat these together as ONE assertion---the claim the source sentence makes. Judge whether that whole claim is covered; do NOT just match the subject/relation/value strings as separate slots. The source sentence tells you what the claim actually means (the raw subject/relation/value may be terse or awkwardly worded---rely on the source sentence to understand the true assertion). \par

Decide COVERED only when the SYNTHETIC TEXT would let a reader RECOVER THE EXACT CLAIM. Apply these rules strictly: \par

COVERED requires ALL of: \par
1. The specific value (the who/what/when/how-much of the claim) is EXPLICITLY present in the synthetic text---not implied, not inferable, not "close". \par
2. It is UNAMBIGUOUSLY ATTRIBUTED to the right entity/attribute, so a reader can tell this value belongs to THIS claim and not something adjacent. \par
3. Numbers, dates, scores, percentages, quantities, and proper names match EXACTLY. No rounding, no approximation, no unit/precision loss. \par

NOT COVERED (be strict---when in doubt, choose NOT COVERED) if ANY of: \par
- the value is missing, only alluded to, or requires inference/world-knowledge to reconstruct; \par
- ANY loss of precision: a full date rendered as year-only (or vice-versa); an exact number/score/percentage rendered as "several"/"many"/"about"/rounded/a range; a specific award/item/rank/table-row replaced by its general category; \par
- the entity is mentioned but this particular attribute/value is not; \par
- the value only appears inside a bare structured fragment or list with no context tying it to the entity (a value that merely "shows up" among enumerated items does NOT count unless the text clearly asserts it OF the entity); \par
- the attribution is uncertain or could plausibly belong to a different entity. \par

Paraphrasing and reordering are fine ONLY if the exact value and its attribution both survive. \par

SYNTHETIC TEXT: \par
""" \par
\%\%CTX\%\% \par
""" \par

FACTS: \par
\%\%FACTS\%\% \par

Output a JSON array, one object per fact, SAME ORDER, using the Fact number as id: \par
\texttt{[\{"id":1,"covered":true\},}\allowbreak\texttt{\{"id":2,"covered":false\}, ...]}
\end{promptbox}

\section{Source-Availability Funnel}
\label{app:source-availability}
\suppressfloats[t]

To distinguish missing source evidence from omissions introduced by synthesis, we instrument SimpleQA Verified as a source-availability probe against the FineWiki training corpus. Table~\ref{tab:funnel} reports whether the corresponding source article is present and whether the answer is verifiably present in that article. This diagnostic is separate from the downstream training results in the main text.

\begin{table}[t]
\centering\small
\caption{Source-availability funnel for SimpleQA Verified against FineWiki. Gap A contains questions whose source article is not found in the corpus; Gap B contains questions whose article is found but whose target answer is not verifiably present in that article.}
\label{tab:funnel}
\setlength{\tabcolsep}{5pt}
\begin{tabular}{lrr}
\toprule
\textbf{Level} & \textbf{Count} & \textbf{\%} \\
\midrule
L1: All questions & 1{,}000 & 100 \\
L2: Source article in corpus & 776 & 77.6 \\
L3: Answer verifiably in article & 555 & 55.5 \\
\midrule
\textit{Gap A: no article (L1$-$L2)} & 224 & 22.4 \\
\textit{Gap B: article but answer absent (L2$-$L3)} & 221 & 22.1 \\
\bottomrule
\end{tabular}
\end{table}

\section{Benchmark and Evaluation Details}
\label{app:bench}

\paragraph{Factual benchmark protocol.} For SimpleQA, SimpleQA Verified, PopQA, and HotpotQA, we use a 5-shot prompt with the format \texttt{Question: ...\textbackslash n Answer: ...}. We do not use retrieval. Each benchmark is evaluated with the same three-stage pipeline:
\begin{enumerate}[nosep,leftmargin=*]
\item \textbf{Inference}: the model generates an answer under the specified decoding setting.
\item \textbf{Judge}: Qwen2.5-7B-Instruct assigns one of \texttt{CORRECT}, \texttt{INCORRECT}, or \texttt{NOT\_ATTEMPTED}.
\item \textbf{Score}: accuracy is the proportion of examples judged \texttt{CORRECT}; \texttt{INCORRECT} and \texttt{NOT\_ATTEMPTED} are both counted as non-correct.
\end{enumerate}
For pass@1, we use temperature 0 and greedy decoding. For sampled pass@$k$ with $k>1$, we use temperature 0.6 and top-$p=0.95$; a question is counted as correct if any of the $k$ sampled answers is judged \texttt{CORRECT}. Table~\ref{tab:sqa-verified-passk} reports $k\in\{1,8,16,32,64\}$ on SimpleQA Verified.

\paragraph{Judge prompt.}
We use the SimpleQA-style grader template with gold aliases and model predictions as inputs, and constrain the judge to output one of \texttt{CORRECT}, \texttt{INCORRECT}, or \texttt{NOT\_ATTEMPTED}.

\paragraph{Judge audit.} We stratify sampled benchmark-judge decisions by model size, method, and judge label, and the authors manually review 500 examples. The automatic judge agrees with the author review on 493/500 examples (98.6\%).

\paragraph{Bootstrap uncertainty.} Table~\ref{tab:bootstrap-se} reports accuracy with bootstrap standard errors. Paired bootstrap tests show that the 7B AFCS gains over both WRAP and Nemotron-CC are significant on all four factual benchmarks ($p<0.001$).

\begin{table}[!t]
\centering\small
\caption{Factual benchmark accuracy (\%) with bootstrap standard error. SQA and SQA-V denote SimpleQA and SimpleQA Verified.}
\label{tab:bootstrap-se}
\setlength{\tabcolsep}{5pt}
\begin{tabular*}{\textwidth}{@{\extracolsep{\fill}}llcccc}
\toprule
\textbf{Model} & \textbf{Size} & \textbf{SQA} & \textbf{SQA-V} & \textbf{PopQA} & \textbf{Hotpot} \\
\midrule
Base & 7B & $3.63{\pm}0.17$ & $3.95{\pm}0.35$ & $23.06{\pm}0.20$ & $27.78{\pm}0.22$ \\
WRAP & 7B & $3.99{\pm}0.15$ & $4.22{\pm}0.37$ & $25.74{\pm}0.19$ & $27.08{\pm}0.21$ \\
Nemotron-CC & 7B & $3.99{\pm}0.22$ & $4.16{\pm}0.39$ & $26.42{\pm}0.20$ & $27.40{\pm}0.22$ \\
\textbf{AFCS} & \textbf{7B} & $\mathbf{5.63{\pm}0.30}$ & $\mathbf{6.17{\pm}0.33}$ & $\mathbf{29.80{\pm}0.21}$ & $\mathbf{29.82{\pm}0.25}$ \\
\midrule
Base & 32B & $5.94{\pm}0.35$ & $5.78{\pm}0.36$ & $29.17{\pm}0.21$ & $34.50{\pm}0.24$ \\
WRAP & 32B & $7.10{\pm}0.33$ & $7.20{\pm}0.38$ & $32.46{\pm}0.21$ & $35.07{\pm}0.25$ \\
Nemotron-CC & 32B & $6.82{\pm}0.35$ & $7.24{\pm}0.44$ & $33.01{\pm}0.18$ & $34.60{\pm}0.24$ \\
\textbf{AFCS} & \textbf{32B} & $\mathbf{9.11{\pm}0.28}$ & $\mathbf{9.08{\pm}0.41}$ & $\mathbf{37.22{\pm}0.20}$ & $\mathbf{35.40{\pm}0.25}$ \\
\bottomrule
\end{tabular*}
\end{table}

\paragraph{General benchmark suite.} The 12-task mid-training suite comprises ARC \citep{clark2018arc}, MMLU-Redux \citep{gema2024mmlu}, SQuAD v2 \citep{rajpurkar2018know}, DROP \citep{dua2019drop}, OpenBookQA \citep{mihaylov2018openbookqa}, CommonsenseQA \citep{talmor2019commonsenseqa}, WinoGrande \citep{sakaguchi2020winogrande}, PIQA \citep{bisk2020piqa}, HellaSwag \citep{zellers2019hellaswag}, GSM8K \citep{cobbe2021gsm8k}, WikiTableQuestions \citep{pasupat2015wikitablequestions}, and TriviaQA \citep{joshi2017triviaqa}. Scores follow the evaluation harness used in the OLMo-3 mid-training comparison and are reported on a 0--100 scale; SQuAD v2 and DROP use their F1-based scores.

\paragraph{Compute.} The 7B pretraining runs use 64 H20 GPUs; one 20B-token run takes $\sim$36 hours. Each 32B run uses 256 H20 GPUs for $\sim$48 hours; total compute is $\sim$5$\times$ a 7B run. Synthesis is run separately on an inference cluster.

\section{FineWiki Raw Controls}
\label{app:raw-baseline}

\begin{table}[H]
\centering\small
\caption{AFCS versus continued pretraining on raw FineWiki. SQA and SQA-V denote SimpleQA and SimpleQA Verified; values are judge-based accuracies (\%).}
\label{tab:raw-controls}
\setlength{\tabcolsep}{5pt}
\begin{tabular*}{\textwidth}{@{\extracolsep{\fill}}llrrrr}
\toprule
\textbf{Model} & \textbf{Size} & \textbf{SQA} & \textbf{SQA-V} & \textbf{PopQA} & \textbf{Hotpot} \\
\midrule
FineWiki Raw & 7B & 3.90 & 4.90 & 25.23 & 28.26 \\
\textbf{AFCS} & \textbf{7B} & \textbf{5.63} & \textbf{6.17} & \textbf{29.80} & \textbf{29.82} \\
\midrule
FineWiki Raw & 32B & 6.97 & 7.26 & 31.96 & 35.29 \\
\textbf{AFCS} & \textbf{32B} & \textbf{9.11} & \textbf{9.08} & \textbf{37.22} & \textbf{35.40} \\
\bottomrule
\end{tabular*}
\end{table}

Table~\ref{tab:raw-controls} directly compares AFCS with continued pretraining on the raw FineWiki source corpus at both scales. AFCS improves all four factual benchmarks; at 32B, the HotpotQA margin is 0.11 points (35.40 vs.\ 35.29).

\section{Mid-Training Evaluation Benchmarks}
\label{app:midtrain-breakdown}

\begin{table}[H]
\centering\small
\caption{Per-benchmark breakdown for 100B-token mid-training integration. Mid. denotes the 100B general mid-training baseline.}
\label{tab:midtrain-perbenchmark}
\setlength{\tabcolsep}{5pt}
\begin{tabular*}{\textwidth}{@{\extracolsep{\fill}}llrrr}
\toprule
\textbf{Category} & \textbf{Benchmark} & \textbf{Base} & \textbf{Mid.} & \textbf{AFCS} \\
\midrule
\multirow{2}{*}{Gen. Know.}
 & ARC & 77.47 & 84.98 & \textbf{85.24} \\
 & MMLU Redux & 60.30 & 65.30 & \textbf{66.12} \\
\midrule
\multirow{2}{*}{Read. Comp.}
 & SQuAD v2 & 42.27 & 48.59 & \textbf{52.96} \\
 & DROP & 40.51 & \textbf{66.51} & 65.93 \\
\midrule
\multirow{2}{*}{Reasoning}
 & OpenBookQA & 76.40 & \textbf{85.20} & 82.40 \\
 & CSQA & 72.15 & \textbf{74.20} & 73.87 \\
\midrule
\multirow{3}{*}{Lang. Und.}
 & WinoGrande & 53.28 & \textbf{65.11} & 64.72 \\
 & PIQA & 73.99 & 74.54 & \textbf{76.50} \\
 & HellaSwag & 59.83 & 70.15 & \textbf{76.41} \\
\midrule
Math & GSM8K & 38.36 & \textbf{79.53} & 79.08 \\
\midrule
\multirow{2}{*}{Table/Trivia}
 & WikiTableQ & 36.55 & 43.74 & \textbf{45.79} \\
 & TriviaQA & 62.42 & 61.01 & \textbf{62.95} \\
\midrule
\multicolumn{2}{@{}l}{\textbf{Average}} & 57.79 & 68.24 & \textbf{69.33} \\
\bottomrule
\end{tabular*}
\end{table}

The 12 general benchmarks used in Section~\ref{sec:midtrain} cover general knowledge, reading comprehension, reasoning, language understanding, math, table understanding, and trivia-style QA. Table~\ref{tab:midtrain-perbenchmark} reports each benchmark's evaluation-harness score on a 0--100 scale; SQuAD v2 and DROP use F1-based scores.

\end{document}